# Unified Smart Factory Model: A model-based Approach for Integrating Industry 4.0 and Sustainability for Manufacturing Systems

Ishaan Kaushal*[1], Amaresh Chakrabarti[1]

[1] Department of Design and Manufacturing, Indian Institute of Science, Bangalore, India

* Corresponding author. Tel.: +91-9805817595; E-mail address: kaushalishaan@gmail.com; Address: IDeaS Lab, Department of Design and Manufacturing, Indian Institute of Science, Bangalore, India -560012

**Abstract**
This paper presents the Unified Smart Factory Model (USFM), a comprehensive framework designed to translate high-level sustainability goals into measurable factory-level indicators with a systematic information map of manufacturing activities. The manufacturing activities were modelled as set of manufacturing, assembly and auxiliary processes using Object Process Methodology, a Model Based Systems Engineering (MBSE) language. USFM integrates Manufacturing Process and System, Data Process, and Key Performance Indicator (KPI) Selection and Assessment in a single framework. Through a detailed case study of Printed Circuit Board (PCB) assembly factory, the paper demonstrates how environmental sustainability KPIs can be selected, modelled, and mapped to the necessary data, highlighting energy consumption and environmental impact metrics. The model's systematic approach can reduce redundancy, minimize the risk of missing critical information, and enhance data collection. The paper concluded that the USFM bridges the gap between sustainability goals and practical implementation, providing significant benefits for industries specifically SMEs aiming to achieve sustainability targets.

Keywords: Sustainable Manufacturing; MBSE; Information Model; Industry 4.0; LCA

## 1) Introduction

The manufacturing industry uses various resources such as materials and energy to produce goods, but it also generates significant amounts of pollutants, effluents, and solid waste that are released into the environment [1,2]. Several factors, including climate change, stricter government regulations, customer preferences for environmentally friendly products, and the increased rate of resource depletion, have prompted organizations to focus on improving their environmental performance [3–5]. To address these sustainability issues in manufacturing, a new disciplines like sustainable production and sustainable manufacturing have emerged over the past few decades [6,7]. Sustainable manufacturing is defined as the creation of manufactured products using processes that minimize negative environmental impacts, conserve energy and natural resources, and are safe for consumers, communities, and economically viable [7]. By adopting sustainable manufacturing practices, organizations can reduce their environmental footprint while also improving their economic performance through increased efficiency and reduced waste.

The quantification of environmental impacts is a critical step in implementing sustainable manufacturing practices. It enables organizations to make informed decisions about selecting sustainable products, processes, or systems. To achieve this, various sustainability assessment methods are used, including the Dow Jones Sustainability Index (DJSI), Global reporting initiative (GRI), Eco-points, Ecological Footprint and Life Cycle Assessment (LCA) [8]. LCA is the most effective and holistic method for quantifying environmental impacts in manufacturing. LCA considers the entire life cycle of a product or process, from raw material extraction to disposal, and provides a comprehensive evaluation of its environmental impacts [9]. However, accurate estimations of environmental impacts require reliable and robust data. Organizations need to collect and analyse data on material and energy inputs, emissions, waste generation, and other factors to accurately quantify the environmental impacts of their manufacturing processes. The quality of the data used in LCA, or any other sustainability assessment methods directly impacts the accuracy of the results[10]. In addition to the capturing data, interpreting the results of sustainability assessment methods and using them to improve manufacturing processes is a significant challenge. To address these challenges (i.e., data capture and result interpretation), there is a need to develop standard and detailed manufacturing process models that can help identify areas for improvement and implement changes to reduce their environmental impact [11]. Manufacturing process models provide a consistent and uniform approach for analysing the inputs, processes, and outputs of manufacturing processes [12]. By developing standard and detailed manufacturing process models and using sustainability assessment methods, organizations can improve their environmental performance and contribute to a more sustainable future [10]. There is another significant challenge when it comes to translating sustainability goals and indicators into manufacturing-level data requirements. The sustainability goals are broad, and to achieve them, a detailed roadmap must be laid out that outlines how manufacturing affects these goals, and which elements and processes of manufacturing contribute to sustainability objectives.

The research objectives of this paper are twofold: 1. Standardize mapping of manufacturing systems for performing LCA (i.e., Process modelling, KPI selection and assessment) in SMEs, and 2. To apply the method in a real-world scenario. The first objective was achieved through the development of systematic guidelines based on a top-down approach for performing LCA, with the aim of reducing redundancies and enhancing efficiency to save costs and effort. As part of the guidelines, a procedure for modelling the manufacturing process and system using Object Process Methodology (OPM) was also developed. The second objective was addressed by demonstrating the applicability of the support in the manufacturing line of a PCB assembly factory.

The paper has been organized into different sections to answer the above questions. Section 2 presents the background of methods used in the paper. Section 3 addresses the challenges in the implementation of Digital transformation for sustainability assessment in Manufacturing Systems by presenting a method for KPI selection and Assessment and MBSE based manufacturing process and system representation. Section 4 presents a case study to show the implementation of MBSE based method to represent manufacturing process and system for data collection and how it can be integrated to perform sustainability assessment. Finally, Section 5 and Section 6, presents, conclusions and discussion. Supplementary Material S1 is also added to support section 3.1 (https://doi.org/10.5281/zenodo.17803646).

## 2) *Background:*

### 2.1. Adoption of sustainability and Industry 4.0 into SMEs

Sustainability has become a global priority due to climate change and other environmental challenges. However, adapting sustainability goals and principles for small and medium enterprises (SMEs) at shop floor level is complex. SMEs do recognize the need to address environmental concerns and seek strategies to improve cost-effectiveness and competitiveness while reducing negative impacts [13], yet they often lack awareness of sustainability strategies and practices. SMEs make up the bulk of enterprises in most nations, making them important for economic growth and innovation. Yet, compared to large companies, the literature has paid less attention to how sustainability strategies affect the success of SMEs [14].

The emergence of Industry 4.0 in recent times has presented significant opportunities for the improvement of the manufacturing sector. It is not only providing economic benefits but also presenting solutions for achieving environmental sustainability [15]. The synergy between Industry 4.0 and sustainability has the potential to revolutionize global production systems. [16], [17]. However, SMEs face significant barriers to adopting Industry 4.0, including limited expertise, capital, and infrastructure. Unlike larger corporations, SMEs are still in the early stages of implementation and addressing these obstacles is crucial to accelerating their adoption and unlocking Industry 4.0's potential. [14,18].

Industry 4.0 and sustainability are multidisciplinary concepts and are, therefore, challenging to grasp [1,19]. Researchers have simplified the concepts by breaking down the whole Industry 4.0 concept into individual elements, such as physical systems, digital systems, communication, and the cloud, and presented conceptual models for the implementation of Industry 4.0 concepts. Researchers have proposed models to assess readiness of manufacturing companies for Industry 4.0 and such models can help in planning the transition [20–22]. There is still no consensus on such models but still these models can assist in road mapping the journey towards Industry 4.0. In addition to that, SMEs also face multiple challenges like digitally skilled workforce shortage, financial challenges, lack of collaboration, and technological challenges like mixed equipment with varied levels of automation [23,24].

Sustainability decisions require robust data, and Industry 4.0 offers the capability to collect and analyse data across the value chain for informed decision-making [25]. However, challenges remain in identifying what data to collect and from which resources. High-fidelity models require mapping appropriate data sources without redundancy, adding complexity at both micro and macro levels [26] [27]. The adoption of Industry 4.0 also introduces interdisciplinary challenges, such as managing diverse manufacturing processes, systems, products, and automation levels. [28], [29] [30]. All such issues can lead to a lot of redundancy in the modelled system and modellers can lose track of mapping the deliverables to data requirements and data sources. To overcome these challenges, lessons can be drawn from other complex engineering domains like aerospace, defence, and space technologies, where complexity is effectively managed using Model-Based Systems Engineering (MBSE). MBSE provides a structured approach to handle interconnected systems and processes, ensuring that the goals are met while maintaining clarity and reducing ambiguity.

*2.2. Object-Process Methodology (OPM)*

OPM is Model based systems engineering (MBSE) language to system development and specification that integrates the major system aspects, including function, structure, and behaviour, into a single graphical and textual model. Model-based systems engineering (MBSE) is the formalized application of modelling to support system requirements, design, analysis, verification and validation activities beginning in the conceptual design phase and continuing throughout development and later life cycle phases [31]. MBSE based modelling improves complexity management, brings traceability to changes and shareability across teams in a consistent manner (Friedenthal, Griego, and Sampson 2007). The OPM model depicts the two fundamental aspects of a system: its objects and processes, and it provides a simplistic approach to model systems using stateful objects, processes, and links among them [33]. ISO has adopted and published OPM as ISO 19450—Automation systems and integration—Object-Process Methodology (ISO 2015.). OPM represents the structure and behaviour of a system using objects and processes, respectively. Objects represent the things that exist in a system, and processes transform those objects by creating, consuming, or changing their states [35]. Structural links connect objects to objects or processes to processes and represent the static structure of the system. Procedural links connect objects to processes and represent the dynamic behaviour of the system [35]. By combining these elements and showing how they relate to each other, a systems engineer can describe complex systems and their functions and behaviours [36]. As the manufacturing systems and supply chains are getting highly complex [29], such a precise model reduces the complicatedness of the system representation and hence better understanding and communication across the multiple stakeholders. The other significant advantages of modelling systems with OPM are bimodal graphical-textual representation and its built-in complexity management mechanisms of in-zooming and unfolding of a single type of diagram.

*2.2.1 Main ISO 19450-compliant OPM Symbols:*

OPM symbols used for manufacturing system modelling in this paper are shown in Figure 1. OPM represent both stateful objects and processes and explicitly shows the connections between them using links. Processes create, consume, or transform objects from one state to another. An object is a thing that exists or can exist physically or informationally/ conceptually/ logically. For example, Bill of Materials (BOM) is an informational object while CNC machine is physical object. Physical objects are represented by a shaded rectangle while informational objects are represented by flat rectangle. A state is a possible situation at which an object can be, or a value it can assume, for some positive amount of time. For example, for CNC machine three states can be off, idle and in-process running. States are part of the object and represented as chamfered rectangle inside the object rectangle. A process is a thing that transforms an object physically or informationally. For example, inventory management is an informational process while machining is a physical process. Physical processes are represented by shaded ovals while informational processes are represented by flat ovals. All environmental objects and processes are represented by dashed outlines.

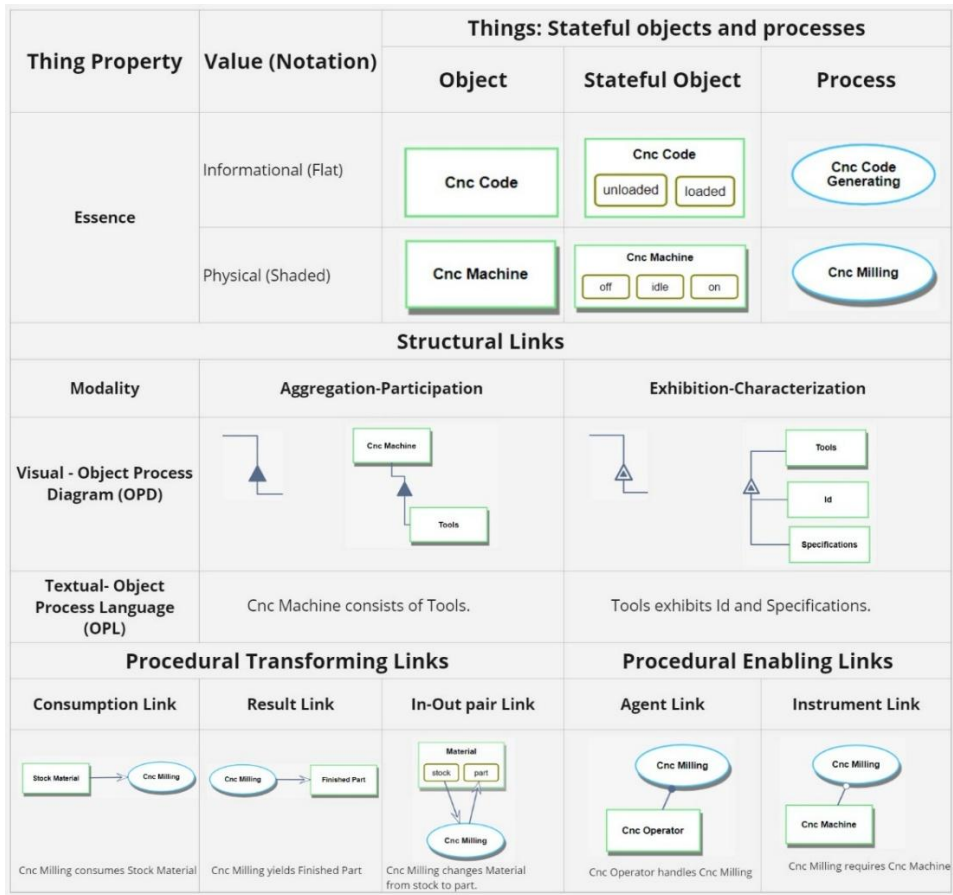

Figure 1: OPM symbols (adapted from [35])

*2.3. SMS model*

Smart Manufacturing Systems (SMSs) are becoming increasingly complex and diverse, making it challenging to map and collect accurate data. To address this issue, a conceptual model for SMSs (SMS model) has been proposed and expanded to facilitate data collection at the factory level [10,37]. The model captures the details of the manufacturing system and process/activity levels, providing valuable insights into the manufacturing activities.

At the manufacturing system level, the model collects information on the type of products being manufactured, the complexity of the manufacturing system, the automation level, building services, inventory and production control strategies, and waste segregation and disposal strategies. The SMS model is an Input-Output process model at the process/activity level, with inputs and outputs categorized as Material Objects (MO), Energy (E), Information (I), Equipment (Eq), Human (H), and Environment (Env). This approach allows for mapping the entire manufacturing system, including Processes/Activities, location, time, and modes of operation.

The model further details inputs and outputs (Material, energy, Information), Equipment, Humans, and Environment of/for each process/activity, down to the attribute level. Inputs and outputs comprise multiple objects (O) and are characterized by one or more attributes (A). Real-time assessment of these attributes is critical for improving manufacturing systems and provides insight into what needs to be measured, from where, and how. Depending on the type of study, attributes can be identified, and a plan for their measurement can be developed. The model has been successfully used to model a shoe factory to collect data for performing Life cycle assessment and the model's efficacy in collecting data comprehensively has been demonstrated [10].

### 3) *Proposed Research Method:*

The paper proposes two methods: a conceptual model for representing manufacturing processes and systems using OPM, an MBSE language; and an integrated model for selection and assessment of Sustainability KPIs. The importance of KPI selection and assessment framework lies in the fact that it is hard for enterprises to translate higher level goals to the process level. Business leaders in companies make goals based on legislations, customer requirements etc. at a higher level like net zero goals. There needs to be a systematic way of translating these goals to process level metrics or attributes. In this paper, a step-by-step methodology is presented for selecting sustainability KPIs and how they translate to data requirement at the factory floor.

#### *3.1. Conceptual model for representing manufacturing process and system using OPM.*

The SMS model is generic and can help collect data for various types of performance assessment of Manufacturing Systems, such as economic assessment, environmental assessment etc [37]. The performance assessment goals decide the Key Performance Indicators (KPIs), and the data required to measure those KPIs can be mapped from the factory floor using the SMS model, as shown in Fig. 2. The SMS model helps in creating a detailed map for each station irrespective of who is collecting the data, and hence helps in improving coverage, reducing uncertainty, and increasing the robustness of the data collection process and the efficiency of data collection using SMS model was validated in our previous work (Kaushal and Chakrabarti 2022), however, there were formalization and complexity management issues with the existing SMS model which can lead to data management issues in larger systems. Standardization is essential for communication among teams and stakeholders and for executing any changes in the system. Modern manufacturing systems are complex, and mapping the entire system becomes highly complex, especially with implementing Industry 4.0 and emerging performance measurement indicators. A detailed map of the manufacturing system made using the SMS model looks complex, and information retrieval and gaining insights from it becomes challenging.

This paper presents a modified and improved MBSE-based SMS model that is formalized using OPM. This model aims to address the issues of formalization and complexity management issues with the existing SMS model and provide a standardized and simplified approach to modelling manufacturing systems. The existing SMS model is formalized using OPM and is presented in the Figure 2 with its language description. The model is created using an open-source tool called OPCAT [38].

The first step of mapping a manufacturing system remains the same as in SMS model, where the system level details are modelled in an 8-step process. The process considers the type and complexity of the product, complexity of the manufacturing system, technical building services, manufacturing system type, automation level, inventory control strategies, production control, and waste segregation/disposal. This information provides a high-level view of the factory's manufacturing system, control strategies, and related goals. For more detailed information, individual processes and their material objects, equipment, energy, human, information, and environment are modelled using the SMS model.

The 'process' here represents a manufacturing process. Each manufacturing process has a specific location, start and end times and accompanied by a set of input and output objects (Input and Output Material Objects, Input and Output Energy and Information Feedback), conditions (input Information), requirements (Equipment) and agents (Human) along with a process environment. The shaded objects or processes are physical in nature like a CNC machine or a machining process while there is no shading for informational objects and processes. Informational objects can be like an algorithm or CNC code, and informational processes can be like phone calling, executing an algorithm etc. The distinction is important as any modern smart manufacturing system consists of both physical and informational entities. The concept of smartness is realized only when physical entities are layered with informational entities.

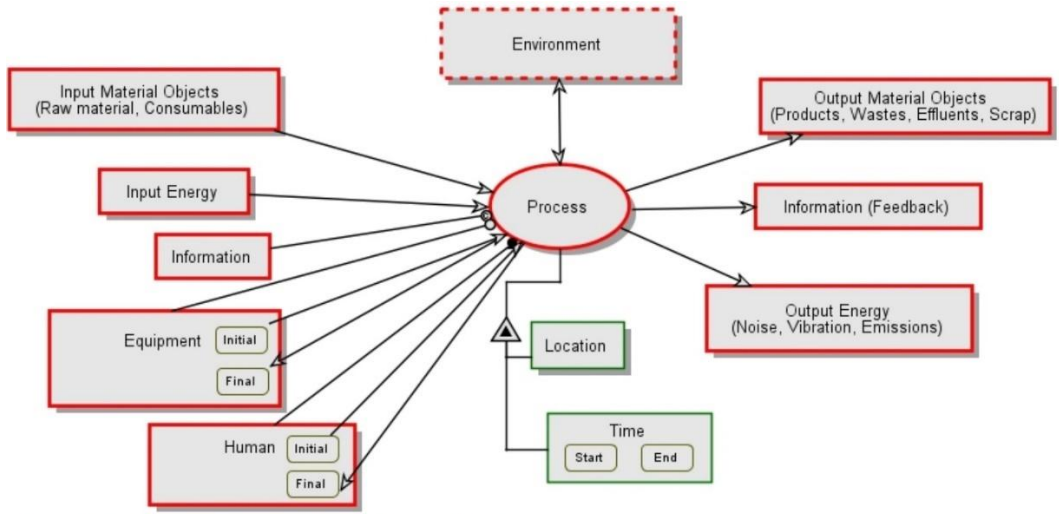

Input Material Objects (Raw material, Consumables) is physical.
Input Energy is physical.
Equipment is physical.
Equipment can be Initial or Final.
Human is physical.
Human can be Initial or Final.
Human handles Process.
Output Material Objects (Products, Wastes, Effluents, Scrap) is physical.
Output Energy (Noise, Vibration, Emissions) is physical.
Environment is environmental and physical.
Process is physical.
Process exhibits Location and Time.
Location is physical.
Time can be Start or End.
Process occurs if Information is in existent.
Process requires Equipment.
Process affects Environment.
Process changes Human from Initial to Final and Equipment from Initial to Final.
Process consumes Input Energy and Input Material Objects (Raw material, Consumables).
Process yields Output Energy (Noise, Vibration, Emissions), Information (Feedback), and Output Material Objects (Products, Wastes, Effluents, Scrap).

Figure 2: OPM-SMS model

### 3.1.1 Process

A process under consideration can be classified into one of nine distinct categories of manufacturing and supporting processes: Primary Shaping, Forming, Separating, Joining, Coating/Finishing, Material Property Change, Auxiliary Processes, Material Handling and Storage, and Testing and Inspection [37,39]. This classification system enables effective categorization and analysis of the industrial processes. This is represented using OPM added in Supplementary Material S1. The Process can be seen in detail by in-zooming the Process in Figure 2. This is a way of managing the complexity of manufacturing systems.

### 3.1.2 Input Material Objects

Input Material Objects are classified into six categories, namely raw material, parts, sub-assemblies, assemblies, interfaces, and consumables [40] added in Supplementary Material S1. The categorization helps in creating better information models. Each of these objects can have attributes ranging from ID, Material type, state, geometry, weight, cost, environmental footprint, prior processing etc. Attributes are informational objects hence they are depicted with rectangular box without shading (Figure 4).

### 3.1.3 Output Material Objects

Output Material Objects are the desired outputs (such as Part, Subassembly or Assembly) of a process and residuals generated during the process. The only difference from Input material objects is the addition of residuals and is added in Supplementary Material S1. The attributes related to parts, assemblies and sub-assemblies are similar to the input ones with addition of residual attributes like the type of residuals produced, their disposal methods, their state (i.e., solid, liquid, or gas), Environmental Impact etc.

3.1.4 *Equipment*

Multiple equipment are used on the factory floor, which broadly can be classified in four categories namely, Machines, Hand tools, Material Handling and Storage, and Computer Systems, added in Supplementary Material S1 as. The attributes associated with these include the following: Name, automation level, type, capacity, manning level, size/footprint, power requirements, tool changes, maintenance schedules, precision, parameters, costs, and environmental impact. In the OPM, Equipment is shown with two states: initial and final; these are to represent the machine states. The connection between an equipment and a process is through an instrument link, which is a procedural enabling link as shown in Figure 2. The categorization of equipment is important to have an estimate of all the requirements for managing these assets.

3.1.5 *Human*

Humans are an important part of any manufacturing system as most of the manufacturing systems in real life are not fully automated. The OPM description of humans are added in Supplementary Material S1. Humans (Operators) conduct various tasks like assembly, material handling, packaging, operating machines etc. Each Operator is associated with one or multiple tasks. The attributes associated with an Operator can be duration of task, number of cycles, shift duration, associated equipment, shop floor accidents and failures, loads handled, grips, movements (involving a part of the body) and ambient working conditions. These attributes are essential for monitoring factory operations as well as the operators' health and safety. Human is linked to the process through an agent link, which is a procedural enabling link as shown in Figure 2.

3.1.6 *Energy*

Energy can be broadly classified into two categories, namely Input and Output Energy. Input Energy can be further subcategorized as energy obtained from an external source and energy generated by humans. The external source of energy can be further classified as electrical, pneumatic, hydraulic, chemical, and other forms. On the other hand, Output Energy can be classified as Energy waste and Recovered Energy. Waste Energy can be further classified into heat, noise, vibrations, and other forms, whereas recovered energy can be classified as recovered heat, steam, electricity through co-generation, and other forms. Each of these categories can have additional attributes such as the amount of energy, energy efficiency, renewable or non-renewable, related equipment, associated individuals, load distribution, energy expenses, and environmental impacts. The OPM description to capture the energy is added in Supplementary Material S1.

3.1.7 Information

Information can be classified into Input Information and Information (Feedback). Input Information can be further categorized into work instruction for humans, work instructions for machines (Algorithms, codes), equipment handling information and information about the previous processes. Information (Feedback) can be sub-categorized into sensor feedback, human feedback, and information for the next process (quality). Information from the previous and for the next processes are explicitly modelled, since this information has instant utility and must be available explicitly to help make decisions in a short span of time. Input Information object is connected to its Process through a conditional link (Figure 2); this means that the process can only be executed if that information is available. The OPM description of Input Information and Feedback are added in Supplementary Material S1.

3.1.8 *Environment*

The environment can be classified into two main categories, namely the immediate environment within the boundaries of a factory, and the outside environment beyond the factory, added in Supplementary Material S1. Several attributes govern these environments, including temperature, humidity, noise levels, and concentrations of various gases. These attributes play a crucial role in measuring and improving the health, safety and environmental issues on the factory floor. Process and environment are linked through affect link which shows that process affects environment as shown in Figure 2. The dotted boundary of environment represents that it is an environmental object.

*3.2. An integrated model for Sustainability KPI selection and assessment*

Digital transformation of the industrial sector is promising to provide solutions to measure sustainability. However, manufacturers face challenges in implementing digitalization and becoming sustainable. To start with, sustainability should be the goal, and digitalization should follow to achieving sustainability. The sustainability goals are very broad and could vary depending on geographical locations as country wise priorities might be different. Manufacturing companies often feel directionless when looking at these goals and may not know where to start and if pursuing sustainability would lead to improved performance. Numerous sustainability metrics have been proposed in the literature related to manufacturing [41,42]. The metrics, along with the associated data from the factory floor, can lead to a robust sustainability assessment. While these metrics are good, they must be connected to industrial goals to track progress and create strategies to achieve the goals. In addition, not all data are easily accessible and must often be measured; if industries wish to accomplish sustainable goals, sustainability performance must be measured continuously rather than reporting only at the end of each year.

The advent of Industry 4.0 and the digitalization of manufacturing systems present a great opportunity to collect data associated with sustainability [43,44]. However, digitalization comes at a cost and effort, and organisations must have a good understanding of their needs modelled in the form of their goals. Based on the existing studies [45–48] and discussions with SMEs, a step-by-step model for KPI selection and assessment is proposed explained in Table 1. The model consists of a ten-step process starting with defining organizational goals and translating them into KPIs. A map of the factory will be created using OPM-SMS model to understand the current system in detail including information about data sources and the capability of the system to collect data. Using knowledge of the KPIs and current system, metrics are proposed for each KPI. For each metric, associated data sources are mapped, and data is collected through a process consisting of measurement, communication, and storage of data. Collected data is then used to model and analyse the metrics and associated KPIs. Decisions are made based on the KPI performance, which can be implemented at the factory floor through human intervention or by automatic controls. Taking this top-down approach can help in better mapping the factory and hence data collection.

Table 1: KPI Selection and assessment framework

| Step | Description | Details |
|---|---|---|
| I | Define Organizational Goals | Goals may include alignment with SDGs, EU directives, or economic, environmental, and social objectives. |
| II | Develop Key performance indicators (KPIs) [49–52] | 1. **KPI Objective**: Define goals, applications, and target audience. |
| | | 2. **KPI Identification**: Determine relevant KPIs. |
| | | 3. **KPI Application Level**: Define levels (unit, process, line, plant) and dimensions (ecological, environmental, societal). |
| | | 4. **KPI Selection**: Apply methods such as Weighted Index Method, AHP, or DSM. |
| III | Describe KPIs [47,48,50] | Elaborate on the KPIs, their definitions, and importance in alignment with goals. |
| IV | Model the Factory | 1. **Manufacturing System**: Describe products, elements, strategies, complexity, automation level, and system boundaries. |
| | | 2. **Processes**: Outline process categories, transformations, material objects, energy, equipment, human resources, information, environment, and boundary conditions. |
| V | Establish Metrics | Define metrics using KPI knowledge and assess system capabilities to capture data. |
| VI | Collect Data from the Factory | Gather measurements, ensure effective communication, and establish data storage methods.(connects to step IV) |
| VII | Create Simulation Models | Develop simulation models to evaluate potential scenarios. |
| VIII | Model and Analyse | Analyse the system based on organizational goals, defined KPIs, assessment methodologies, and boundary conditions. Data is a critical input. |
| IX | Decide and Interpret | Make decisions based on the analysis and interpret results in the context of organizational goals and objectives. |
| X | Implement Decisions at Factory Level | Implement the selected decisions and control measures to achieve the intended outcomes at the operational level. |

The framework includes modelling the manufacturing system indicated in Step IV and has an associated data process. The modelling of manufacturing system is done using OPM-SMS model which helps in mapping the whole factory by mapping each process and associated objects. This map helps in establishing the data availability and associated objects. The map also helps in establishing the kind of infrastructure that exists and augments the data process in deciding how to collect that data. Data process is a sequence of activities to make decision and implement it through human or equipment controls. The process is represented in the Figure 3. In SMEs not all the processes are automated and hence data collection cannot be completely automated and can have some human inputs. The Data process model also helps in gauging if the data collection is manual or automated and helps in planning accordingly to include both kind of information.

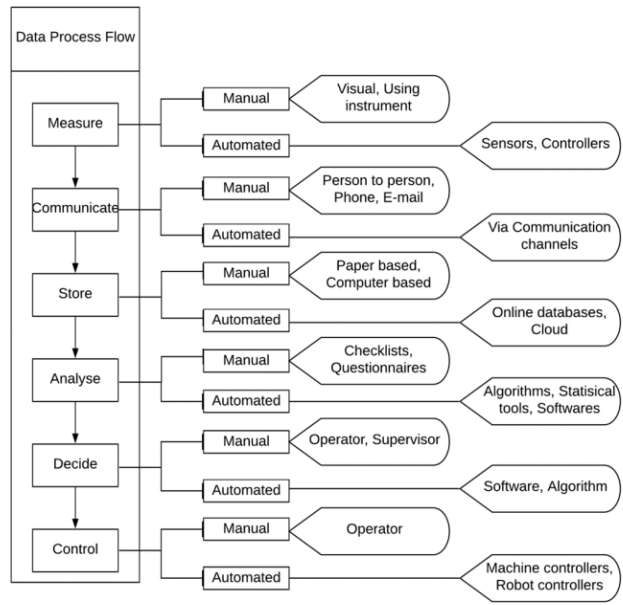

Figure 3: Data Process Model

### 3.3. Unified Smart Factory model (USFM)

Unified Smart Factory model is proposed to represent manufacturing process and system, Data Process and KPI Selection and Assessment method in a single framework. There are three models presented in the above sections. There are overlaps and some distinctions between them but all of them must come together to present a holistic modelling scheme for the smart and sustainable factory. A unified factory model is created by integrating OPM-SMS model, Data Process and KPI Selection and assessment model. The whole model is created using OPM elements and modelled using OPCAT tool. The fundamental idea behind the unified model is that just like manufacturing processes, data processes are also a central element of smart manufacturing systems and must be modelled explicitly. The attributes of each object and process as explained in OPM-SMS model act as the inputs for Data process. Data process is the informational process and takes inputs in the form of information from all physical objects. The information from the physical objects come from their attributes. The Data process is governed by KPI selection and assessment framework from which the information regarding what KPIs to be observed originate. The output of the Data process can be in the form analytics which are shown here as KPI assessment. This holistic model lays down all the elements required and their inter-connections to model a Smart and Sustainable factory. The output from the smart and sustainable factory is not just the physical parts/products but also the insights about the KPIs. The model with the language description is presented in Figure 4.

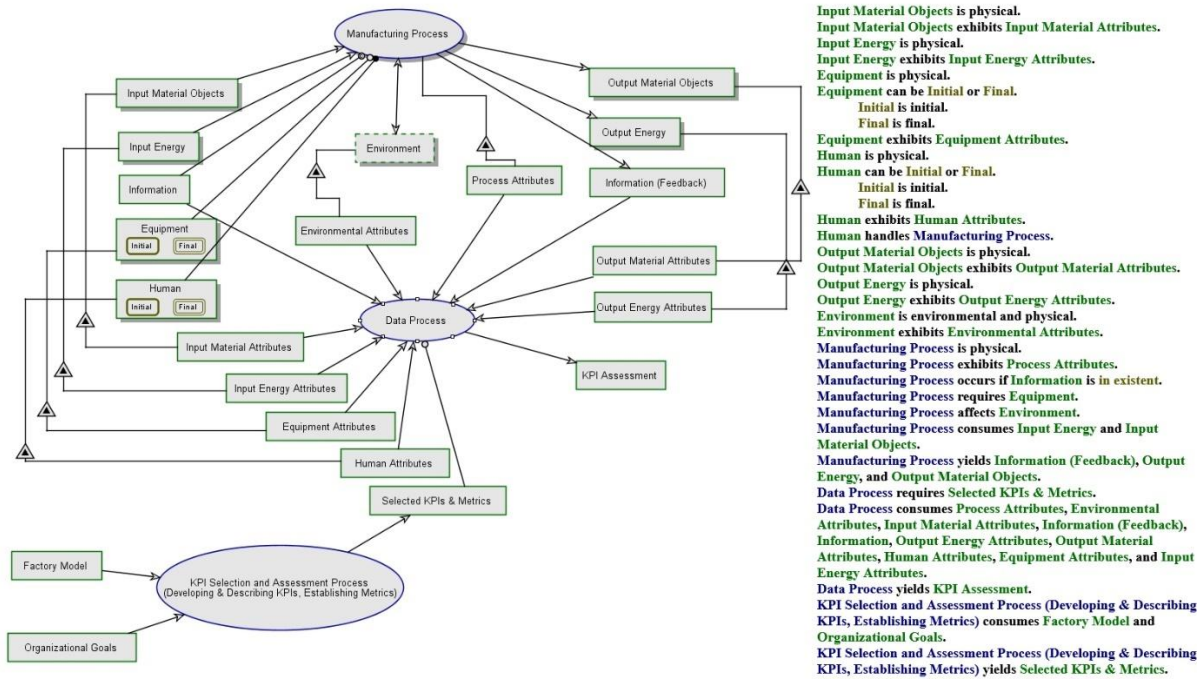

Figure 4: Unified Smart Process Model

*4)   Case Study: Modelling a manufacturing line of PCB assembly factory for performing LCA*

This case study models a real PCB assembly line using the unified smart factory model. The case study is conducted in collaboration with an SME industrial partner that has an automated PCB assembly line. The overall objective of the case study was to examine the environmental performance of the manufacturing system. The case study uses the KPI selection and evaluation framework to select the indicators before narrowing down to the metrics necessary to estimate these KPIs using the OPM-SMS model. The OPM-SMS model is then used to map the factory floor and identify the data sources and associated data required to assess the defined metrics. The full step-by-step methodology is presented below:

Step I: Define organizational goals: Calculating Environmental impacts associated with PCB manufacturing.

Step II: Develop Key performance indicators (KPIs)

1. KPI objective (Goals, applications, audience): The goals of the KPIs are to calculate environmental impacts of producing PCBs at global level. The application of such information is important for the manufacturers, designers, and policy makers in order to decide strategies to reduce the global footprint of such activities. The audience for this study is the manufacturer.
2. KPI identification: Life cycle assessment is the most prominent method to assess environmental impacts associated with products, processes, and services. Using literature on Environmental impact assessment following environmental impact categories were identified: Climate change, Water Depletion, global warming potential, abiotic depletion potential, acidification potential, eutrophication potential, ozone-layer depletion potential, and photochemical oxidation potential. Apart from LCA, Energy consumption, and water consumption were also identified.
3. KPI application level (unit, process, line, plant) and dimensions (Eco, Env, Soc): Though the data comes from each process the whole plant is being considered. There are two applications shown one at the life cycle level (Cradle to Gate) and other at the plant level (Gate to Gate). Energy consumption is measured at the unit level and at the line level while other environmental impact categories were at the regional and global level.

4. KPI selection: The selection among the environmental impact categories is made using the literature defining global impact categories [53]. Though all the impact categories are important but for this study we have considered those impacts which fall under the Global impact categories. Energy Consumption is another common metric found in the literature associated with environmental sustainability and hence was selected for this study [15,49,54–56].

Step III: Describe KPIs

1. Climate change: Climate change is measured by the increased atmospheric concentration of greenhouse gases which, in turn, will increase the radiative forcing capacity, leading to an increase in the global mean temperature (°C) [50]. Increased temperature ultimately results in damage to human health and ecosystems. The method is well documented in most of the Life cycle Impact Assessment (LCIA) methods like Recipe, LC-Impact, Impact world +, Traci etc. [57].
2. Production Energy Consumption: The energy consumed by the production activity [42]. Energy consumption of the assembly line is calculated by analysing the energy data associated with each machine. The energy consumption is normalized by per part basis so that it can be communicated unambiguously.

Step IV: Establish metrics and associated data.

Key data for estimating Environmental Impacts and Production Energy Consumption were identified as material consumption and energy consumption for each machine, time a part spends on each machine, associated event at each machine, timestamp, general consumables, and waste generated. Energy, time, timestamp and event data are automatically collected from each machine while material, consumables and wastage data are collected in aggregated form.

Step V: Model the factory

1. Describe Manufacturing system (Products, elements, strategies, complexity, automation level, system boundaries): The PCB assembly line is a part of bigger manufacturing system and includes five processes[58,59]. The baked board are stacked at the loader and loader moves the boards one by one to the Screen-printing machine. After the screen-printing process, the printed board moves via a conveyor to the pick and place machines 1 and 2 sequentially to place SMT components on the board. Now the Printed board with SMT components goes to the reflow oven to solder all the SMT components and complete PCB comes out of the reflow oven. The whole assembly line is shown in the Figure 5. The complete manufacturing system level information is shown in Table 2.

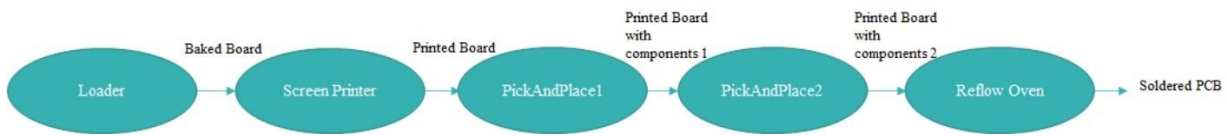

Figure 5: PCB assembly line[59]

Table 2 Manufacturing system level information for selected PCB assembly line

| Type of products manufactured | ISIC Code 2610 - Manufacture of Electronic Components and Boards |
|---|---|
| Complexity | 5 Automated Processes |
| Classification of manufacturing system | Batch Production |
| Automation level | Fully Automated assembly line |
| Production control | Pull system (uses Make-to-Order strategy) |
| Incoming Quality Control | IQC sampling based on ISO:2859 |
| Waste disposal | Waste segregated and sent to third party recycling and disposal units. |

2. Describe Processes: The Process map is created using OPM-SMS model and the KPI information. The metrics associated with each KPI, and the manufacturing system capability were used to identify the data to be collected from the assembly line. Each process shows its inputs and outputs along with equipment and program requirements. Program here refers to the instructions given to the automated machines to carry out the operation and it is connected to the process through the conditional link (process only occurs if program is existing). The energy consumption, time and event attributes are measured from each process to calculate the KPIs associated with each process. Figure 6 and Figure S 9 illustrates the PCB assembly line using OPM-SMS modelling scheme and Object process language (OPL) description.

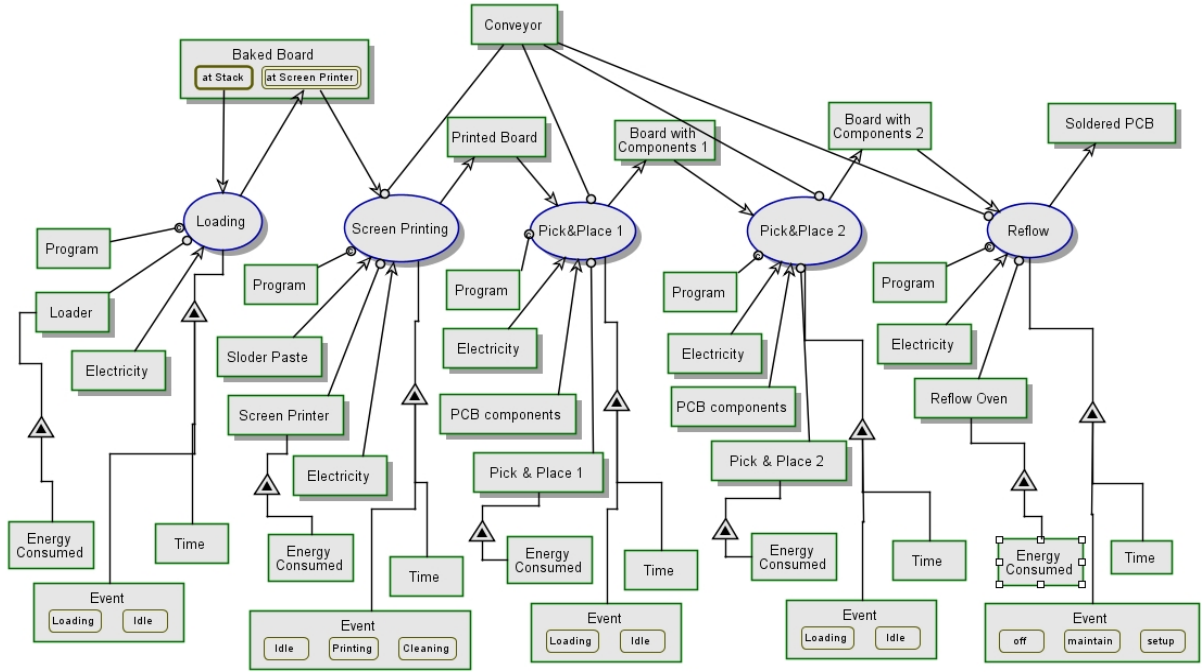

Figure 6: OPM-SMS representation of PCB assembly line

Step VI: Collect data from the factory (Measurement, communication, storage)

The data was collected from the factory floor. Energy, event, and time data was collected automatically from the machines. Material, quality, other consumables, and wastage data was manually collected by visiting the factory floor and meeting the factory supervisor. This data was cumulatively measured and reported and there was no means to collect that automatically at the factory floor. Calculating environmental impacts as indicated in Life Cycle Assessment (LCA) also need data from the whole product life cycle [9]which we did not had access to, and such data was added from the ecoinvent 3.7 database [60].

Step VII: Generate simulation models (optional) In this study no simulation models were generated.

Step VIII: Modelling and Analysis (Analyse)

1. Modelling Energy Consumption: Timestamp data, event data, energy data and time data for each of the 4 machines was automatically collected. The major task in analysis the energy data was to calculate number of PCBs assembled and they were estimated using event data for screen-printing machine. Using these information, average energy consumption for each part throughout the shift was calculated. The total number of parts made in a 14-hour shift were 258 and energy consumed for assembling one PCB was calculated to be 0.47 KWh.
2. Modelling LCA: The goal of the study was to calculate Environmental impact caused by 1 Kg of PCB manufactured. Assembled set of 5 PCBs with equivalent weight of 1 Kg were selected as the functional unit. Product System and system boundary were chosen for the study by considering "Production of Raw material and PCB parts as the background system" and "PCB assembly as the foreground system.". The background system was modelled using EcoInvent 3.7 database in openLCA 1.10.3 software [61]. The foreground system was modelled using the OPM-SMS model. The OPM-SMS model acted as a template to collect information from the factory floor through factory instrumentation, observations, calculations, and inputs from supervisors and operators at the factory floor. ReCiPe 2016 (Hierarchist) midpoint was used as Life cycle impact assessment (LCIA) methodology and Mineral resource scarcity, Global warming, Fossil resource scarcity, Stratospheric ozone depletion, and Water consumption were selected for manufacturing company needs. Inventory data per functional unit is shown in Table 3 and environmental impact scores are reported in Table 4.

Table 3  Life cycle inventory for 1 Kg of PCB assembled.

| Input Flows | | |
|---|---|---|
| Flow | Amount | Unit |
| acrylic varnish, without water, in 87.5% solution state | 0.0078 | kg |
| aluminium, wrought alloy | 0.16 | kg |
| capacitor, electrolyte type, < 2cm height | 0.022 | kg |
| capacitor, electrolyte type, > 2cm height | 0.037 | kg |
| capacitor, film type, for through-hole mounting | 0.076 | kg |
| capacitor, tantalum-, for through-hole mounting | 0.0125 | kg |
| diode, glass-, for through-hole mounting | 0.006 | kg |
| electricity, medium voltage | 5.45 | kWh |
| electronic component factory | 1.15E-09 | Item(s) |
| ethanol, without water, in 99.7% solution state, from ethylene | 0.0013 | kg |
| inductor, ring core choke type | 0.33 | kg |
| integrated circuit, logic type | 0.018 | kg |
| isopropanol | 0.008 | kg |
| mounting, through-hole technology, Pb-containing solder | 0.058 | m2 |
| printed wiring board, for through-hole mounting, Pb containing surface | 0.058 | m2 |
| resistor, metal film type, through-hole mounting | 0.02 | kg |
| section bar extrusion, aluminium | 0.159 | kg |
| transformer, high voltage use | 0.096 | kg |
| transformer, low voltage use | 0.031 | kg |
| transistor, wired, big size, through-hole mounting | 0.026 | kg |
| transistor, wired, small size, through-hole mounting | 0.0009 | kg |
| Chemical waste, unspecified | -0.0005 | kg |
| Printed circuitboards waste | -0.0005 | kg |
| electricity, medium voltage | 2.338 | kWh |
| Output Flows | | |
| PCB_Assembly | 1 | kg |
| used printed wiring boards | 0.0203 | kg |

Table 4: Environmental impact scores

| Name | Impact Value | Unit |
|---|---|---|
| Mineral resource scarcity | 4.37104 | kg Cu eq |
| Global warming | 89.21112 | kg CO2 eq |
| Fossil resource scarcity | 22.95837 | kg oil eq |
| Stratospheric ozone depletion | 4.63E-05 | kg CFC11 eq |
| Water consumption | 0.9449 | m3 |

Step IX: Decision and Interpretation (Decide)

The decisions for improvement can be made based on such information obtained in Step VIII. In case of energy consumption if it goes above a certain threshold red flag can be generated to identify the root causes and improvements can be planned. In case of LCA, a continuous check on environmental impacts is done to have track of the impacts associated with manufacturing. It is important as companies may have to pay penalty if they exceed their emission limits or there might be issues due to more impacts associated with product as compared to their competitors. The data further can be used to create reports like Environment Product Declarations to improve the transparency regarding product sustainability. These decisions were not directly implemented at the factory floor, hence in this study this last step is skipped.

### 5) *Conclusions*

In conclusion, sustainability goals are broadly defined, and considerable effort is required in translating higher level goals to measurables at factory level. Unified Smart Factory Model (USFM) was proposed, which is a step-by-step top-down methodology to map goals of an organization to data at the factory level. By integrating KPI modelling, assessment frameworks, and the OPM-SMS model, USFM enables seamless mapping of high-level sustainability objectives to measurable indicators and corresponding data requirements shown in Figure 4. In the case study, KPI modelling, and assessment were presented to demonstrate how KPIs related to environmental sustainability can be selected and modelled and then mapped to the data required. The data then were mapped from the manufacturing processes using the OPM-SMS model. The mapping helped in identifying the processes, equipment, and associated data. Energy related data was automatically collected and was represented in Figure 6, while LCA modelling related data was a mix of manually and automatically collected data reported in Table 3. Energy consumption modelling showed that the total amount of energy consumed per PCB was 0.5 kWh. Out of all the stations, it was found that energy consumption at the reflow oven was the highest, contributing to about 90% of the total energy consumption. More focus should be paid to such machines that are critical in terms of the KPIs being measured, and plan for better optimisation of resources. Environmental impact scores for mineral resource scarcity, global warming, fossil resource scarcity, stratospheric ozone depletion and water consumption were reported. Such information can be used both for reporting and improving environmental performance of the system. Environmental impact information can be used to compare with the best practices and plan future improvements. Overall the case study showed how to use the method proposed.

### 6) *Discussion*

The USFM model provides a systematic framework for translating broad sustainability objectives into practical actions by explicitly modelling manufacturing processes, activities, and associated data flows. Through its detailed mapping capabilities, the model reduces redundancy and minimizes the risk of missing critical information, even for less experienced practitioners. By providing a comprehensive view of factory processes, their interconnections, and inputs/outputs, USFM enhances the clarity and completeness of data collection process.

One of the key takeaways from the case study is the identification of energy-intensive processes (hotspots), such as the reflow oven, which accounted for a significant portion of energy consumption. The finding like this highlights the importance of focusing on such critical elements to optimize resources and achieve sustainability targets. Furthermore, the integration of environmental impact metrics provides actionable insights for improving performance while facilitating compliance with sustainability reporting standards.

The model due to its basis in MBSE modelling, facilitates unambiguous and effective communication among teams and stakeholders, ensuring a shared understanding of processes, goals, and required actions. This capability is particularly valuable for complex manufacturing environments where cross-departmental collaboration is essential. Further detailed and structured data collection using MBSE based approach, as shown in the comprehensive map in Figure 6, the USFM minimizes ambiguity and ensures accuracy in identifying the relationships between processes, equipment, materials, waste, and energy consumption. This level of detail ensures more reliable sustainability assessments, enabling industries to make informed decisions that align with global standards and improve overall performance. For MSMEs, these capabilities translate into cost-efficient sustainability assessments without the need for third-party consultants, resulting in significant monetary savings.

The scalability of the USFM model positions it as a foundational framework with potential for further refinement and application in diverse manufacturing contexts. It can be adapted to different sustainability goals and seamlessly integrated with emerging Industry 4.0 technologies, such as IoT, AI, and digital twins, to enhance its capabilities.

Future developments could also explore the use of context-specific large language model (LLM)-based agents, built on the USFM framework, to make the model more accessible and user-friendly for SMEs.

The USFM model can provide significant benefits for industries, particularly MSMEs, by enabling improved access to data, factory, and sustainability resources facilitating the creation and implementation of an effective roadmap for improvement within the organization. It can help in optimizing resource usage by pinpointing critical processes and equipment that have the most significant environmental and energy impact in the factories, ensuring targeted improvements. The model's structured data collection approach minimizes redundancies and ensures accuracy and completeness, reducing ambiguity in data collection and ultimately sustainability assessments. Overall, the USFM model bridges the gap between sustainability goals and practical implementation in manufacturing systems.

**Acknowledgements** Special thanks to the management of the PCB assembly factory (name not disclosed) and Prof. Rajesh Sundaresan (Robert Bosch Centre for Cyber-Physical Systems, Indian Institute of Science, Bangalore, India) for giving us their valuable time and data from the factory.

**Funding:** This research did not receive any specific grant from funding agencies in the public, commercial, or not-for-profit sectors.

**Availability of data:** Raw data were generated at PCB manufacturing facility (name not disclosed due to privacy concerns) in Mysore, India. Derived data supporting the findings of this study is included in the manuscript and available from the corresponding author, Ishaan Kaushal, on request.

**Supplementary Material S1**

The Supplementary Material S1 complements the section 3.1 with OPM and OPL description of Process, Input and Output Material Objects, Input and Output Energy, and Information Equipment agents (Human) and Environment. It also compliments the OPM diagram shown in Figure 6.

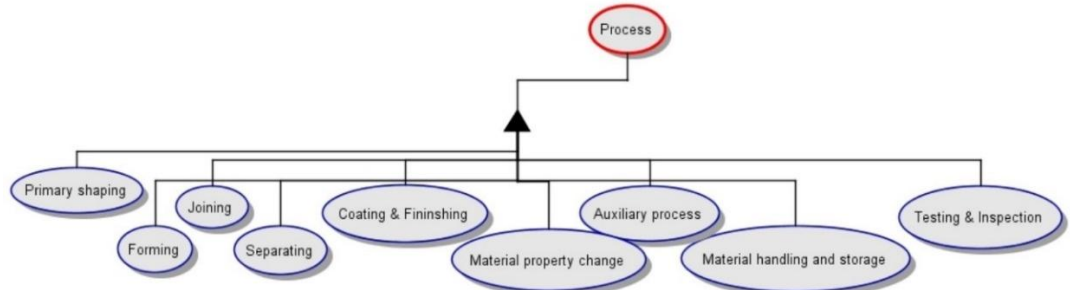

Process is physical.
Process consists of Primary shaping, Joining, Separating, Coating & Fininshing, Material property change, Auxiliary process, Material handling and storage, Testing & Inspection, and Forming.
  Primary shaping is physical.
  Joining is physical.
  Separating is physical.
  Coating & Fininshing is physical.
  Material property change is physical.
  Auxiliary process is physical.
  Material handling and storage is physical.
  Testing & Inspection is physical.
  Forming is physical.

Figure S1: Manufacturing Process classification

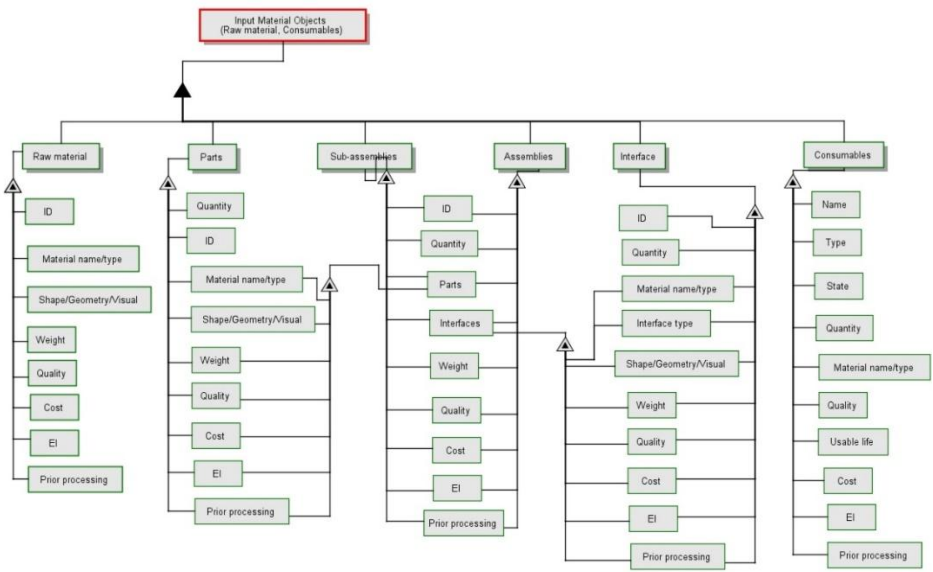

Input Material Objects (Raw material, Consumables) is physical.
Input Material Objects (Raw material, Consumables) consists of Raw material, Parts, Sub-assemblies, Assemblies, Interface, and Consumables.
Input Material Objects (Raw material, Consumables) zooms into Consumables, Interface, Assemblies, Sub-assemblies, Parts, and Raw material.
  Consumables is physical.
  Consumables exhibits Material name/type, Quality, Cost, EI, Prior processing, Quantity, Type, Name, State, and Usable life.
  Interface is physical.
  Interface exhibits ID, Shape/Geometry/Visual, Weight, Material name/type, Quality, Cost, EI, Prior processing, Quantity, and Interface type.
  Assemblies is physical.
  Assemblies exhibits ID, Weight, Quality, Cost, EI, Prior processing, Quantity, Parts, and Interfaces.
    Parts exhibits Shape/Geometry/Visual, Weight, Material name/type, Quality, Cost, EI, and Prior processing.
    Interfaces exhibits Shape/Geometry/Visual, Weight, Material name/type, Quality, Cost, EI, Prior processing, and Interface type.
  Sub-assemblies is physical.
  Sub-assemblies exhibits ID, Weight, Quality, Cost, EI, Prior processing, Quantity, Parts, and Interfaces.
  Parts is physical.
  Parts exhibits ID, Shape/Geometry/Visual, Weight, Material name/type, Quality, Cost, EI, Prior processing, and Quantity.
  Raw material is physical.
  Raw material exhibits ID, Shape/Geometry/Visual, Weight, Material name/type, Quality, Cost, EI, and Prior processing.

Figure S2: Input material objects

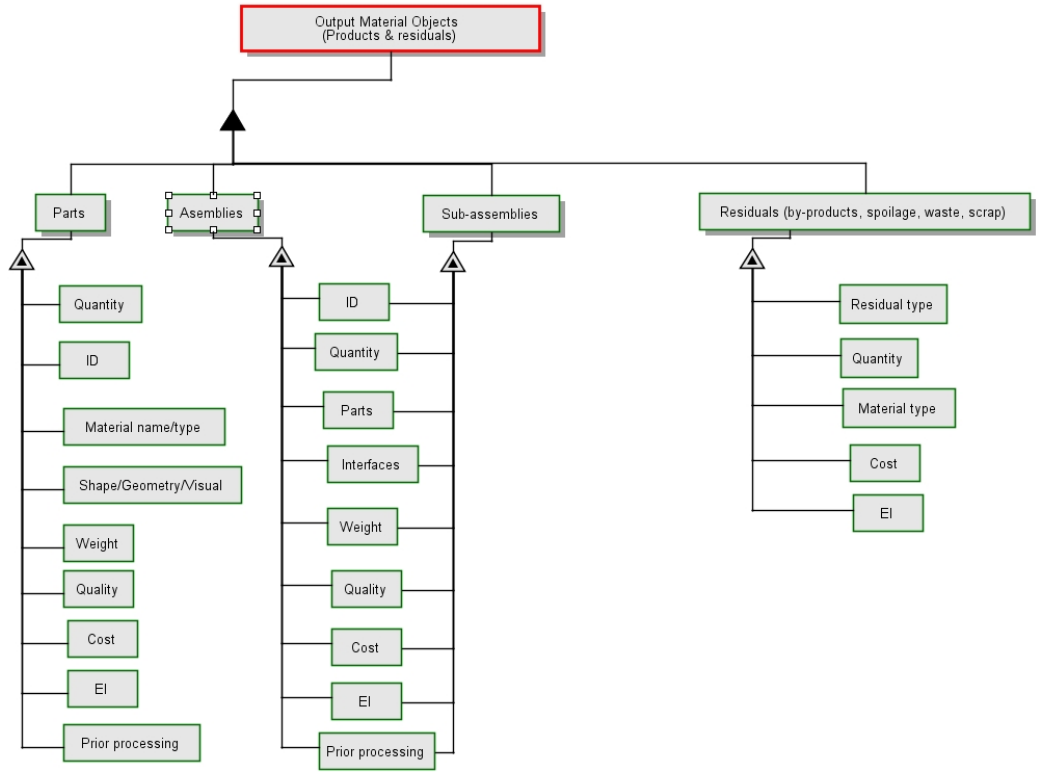

Figure S3: Output material objects

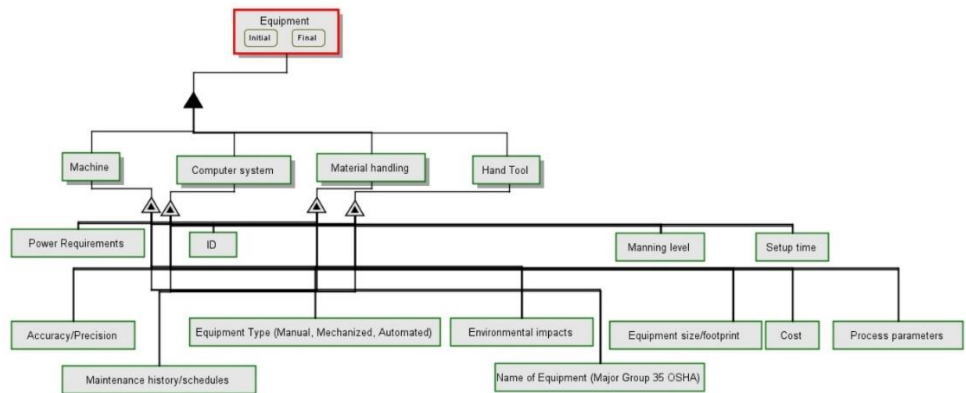

Figure S4: Equipment description

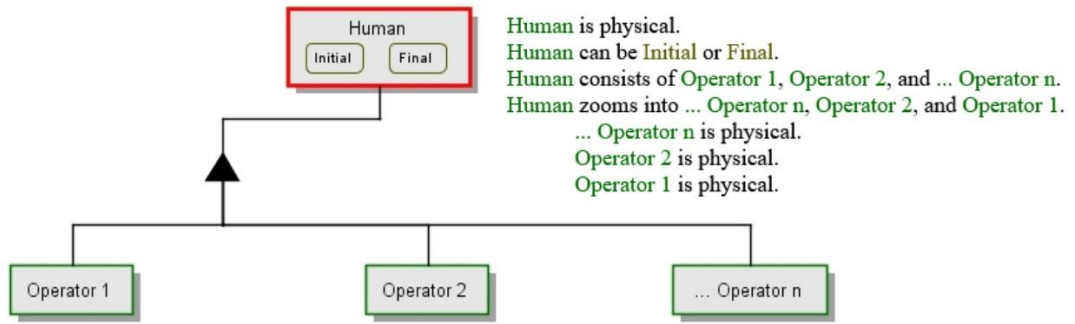

Figure S5: Human description

.

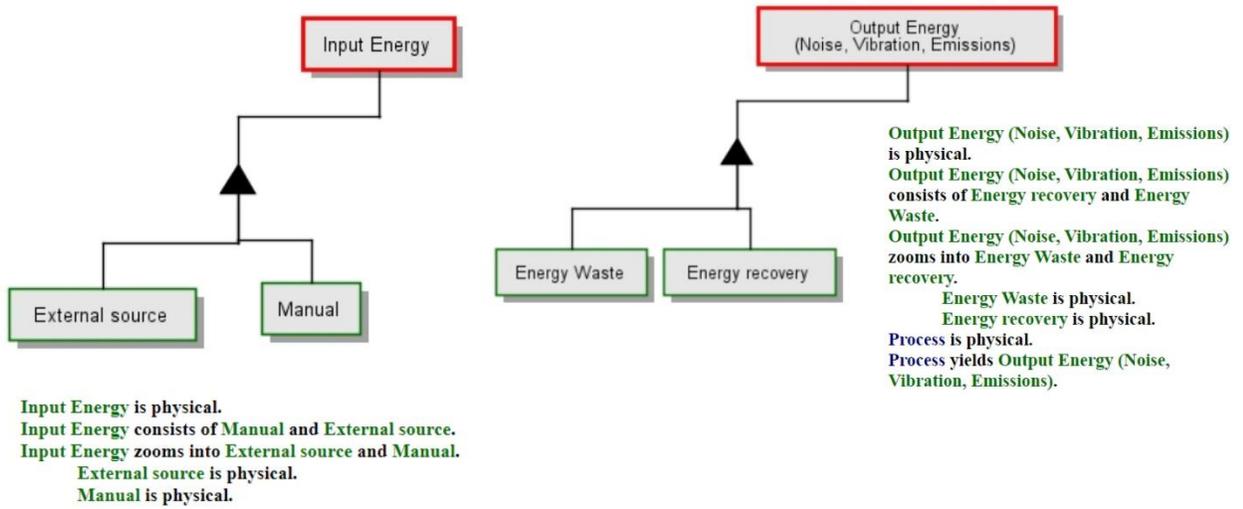

Figure S6: Energy description

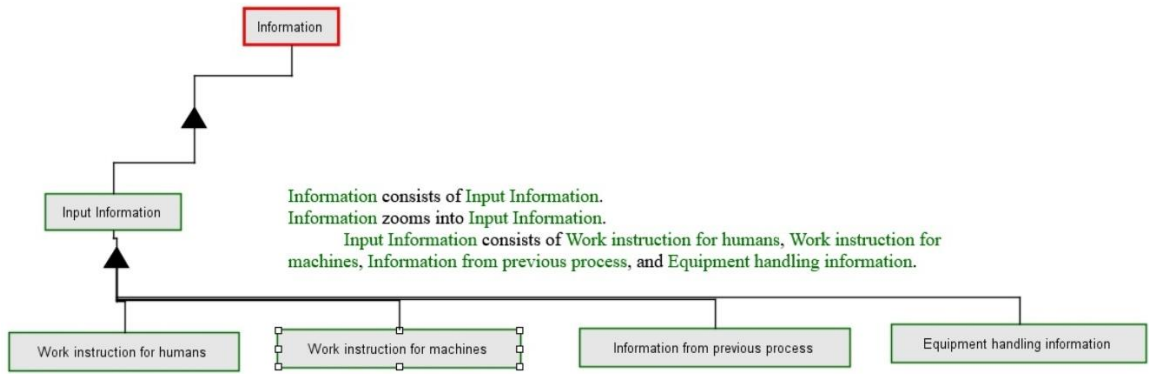

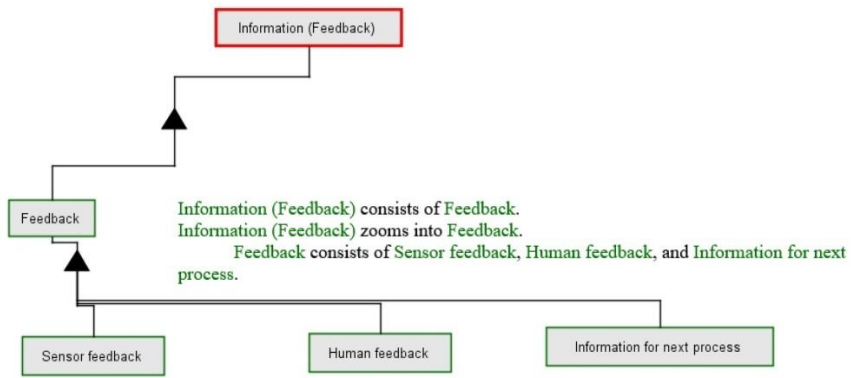

Figure S7: Input information and feedback description

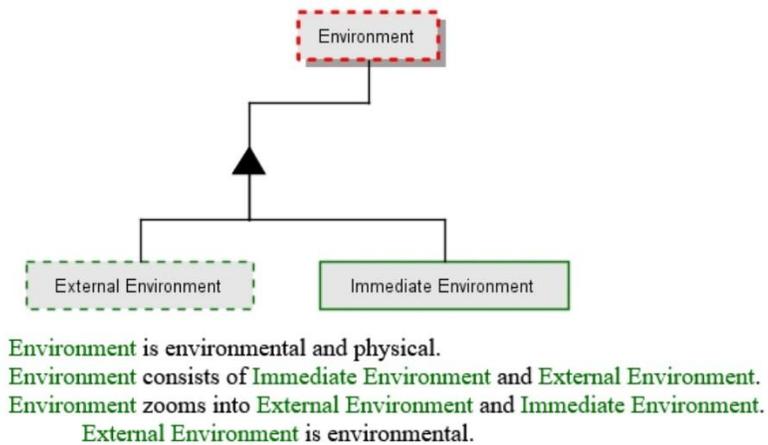

Figure S8: Environment description

Baked Board is physical.
Baked Board can be at Stack or at Screen Printer.
    at Stack is initial.
    at Screen Printer is final.
Printed Board is physical.
Board with Components 1 is physical.
Loader is physical.
Loader exhibits Energy Consumed.
Screen Printer is physical.
Screen Printer exhibits Energy Consumed.
Electricity is physical.
Conveyor is physical.
Pick & Place 1 is physical.
Pick & Place 1 exhibits Energy Consumed.
PCB components is physical.
Sloder Paste is physical.
Soldered PCB is physical.
Pick & Place 2 is physical.
Pick & Place 2 exhibits Energy Consumed.
Reflow Oven is physical.
Reflow Oven exhibits Energy Consumed.
Loading is physical.
Loading exhibits Event and Time.
    Event can be Loading or Idle.
Loading occurs if Program is in existent.
Loading requires Loader.
Loading changes Baked Board from at Stack to at Screen Printer.
Loading consumes Electricity.
Screen Printing is physical.
Screen Printing exhibits Time and Event.
    Event can be Idle, Printing, or Cleaning.
Screen Printing occurs if Program is in existent.
Screen Printing requires Conveyor and Screen Printer.
Screen Printing consumes at Screen Printer Baked Board, Electricity, and Sloder Paste.
Screen Printing yields Printed Board.
Pick&Place 1 is physical.
Pick&Place 1 exhibits Event and Time.
Pick&Place 1 occurs if Program is in existent.
Pick&Place 1 requires Conveyor and Pick & Place 1.
Pick&Place 1 consumes PCB components, Printed Board, and Electricity.
Pick&Place 1 yields Board with Components 1.
Pick&Place 2 is physical.
Pick&Place 2 exhibits Event and Time.
Pick&Place 2 occurs if Program is in existent.
Pick&Place 2 requires Pick & Place 2 and Conveyor.
Pick&Place 2 consumes Board with Components 1, PCB components, and Electricity.
Pick&Place 2 yields Board with Components 2.
Reflow is physical.
Reflow exhibits Event and Time.
    Event can be off, maintain, or setup.
Reflow occurs if Program is in existent.
Reflow requires Reflow Oven and Conveyor.
Reflow consumes Board with Components 2 and Electricity.
Reflow yields Soldered PCB.

Figure S9: OPL representation of PCB assembly line (compliments Figure 6)